\documentclass[runningheads]{llncs}

\usepackage{eccv}

\usepackage{eccvabbrv}

\usepackage{graphicx}
\usepackage{booktabs}
\usepackage{xcolor}
\usepackage{color, colortbl}
\usepackage{bm}

\usepackage[accsupp]{axessibility}  

\usepackage{hyperref}

\usepackage{orcidlink}

\newcommand{\ourMethod}{OPEN}
\definecolor{up}{RGB}{68,169,32}
\definecolor{down}{RGB}{255,0,0}

\newcommand\blfootnote[1]{%
  \begingroup
  \renewcommand\thefootnote{}\footnotetext{#1}%
  \addtocounter{footnote}{-1}%
  \endgroup
}

\begin{document}

\title{OPEN: Object-wise Position Embedding for Multi-view 3D Object Detection} 

\titlerunning{Object-wise Position Embedding for Multi-view 3D Object Detection}

\author{Jinghua Hou\inst{1} 
\and Tong Wang\inst{2}
\and Xiaoqing Ye\inst{2}
\and Zhe Liu\inst{1}
\and Shi Gong\inst{2}
\and Xiao Tan\inst{2}
\and Errui Ding\inst{2}
\and Jingdong Wang\inst{2}
\and Xiang Bai\inst{1\dag}}

\authorrunning{J.~Hou et al.}

\institute{Huazhong University of Science and Technology
\email{\{jhhou,zheliu1994,xbai\}@hust.edu.cn}\\
\and Baidu Inc., China\\
\email{yxq@whu.edu.cn}\\
\email{\{wangtong16,gongshi,tanxiao01,dingerrui,wangjingdong\}@baidu.com}}

\maketitle

\blfootnote{
$\dag$ Corresponding author.
}

\begin{abstract}
Accurate depth information is crucial for enhancing the performance of multi-view 3D object detection. Despite the success of some existing multi-view 3D detectors utilizing pixel-wise depth supervision, they overlook two significant phenomena: 1) the depth supervision obtained from LiDAR points is usually distributed on the surface of the object, 
which is not so friendly to existing DETR-based 3D detectors due to the lack of the depth of 3D object center; 2) for distant objects, fine-grained depth estimation of the whole object is more challenging. Therefore, we argue that the object-wise depth~(or 3D center of the object) is essential for accurate detection. In this paper, we propose a new multi-view 3D object detector named {\ourMethod}, whose main idea is to effectively inject object-wise depth information into the network through our proposed object-wise position embedding. Specifically, we first employ an object-wise depth encoder, which takes the pixel-wise depth map as a prior, to accurately estimate the object-wise depth. Then, we utilize the proposed object-wise position embedding to encode the object-wise depth information into the transformer decoder, thereby producing 3D object-aware features for final detection. Extensive experiments verify the effectiveness of our proposed method. Furthermore, {\ourMethod} achieves a new state-of-the-art performance with 64.4\% NDS and 56.7\% mAP on the nuScenes test benchmark. The code is available at \url{https://github.com/AlmoonYsl/OPEN}.

\keywords{Multi-view 3D object detection \and Depth prediction \and Position embedding}
\end{abstract}
\section{Introduction}
\label{sec:intro}

\begin{figure}[t]
\centering
\vspace{-10pt}
\includegraphics[width=0.99\linewidth]{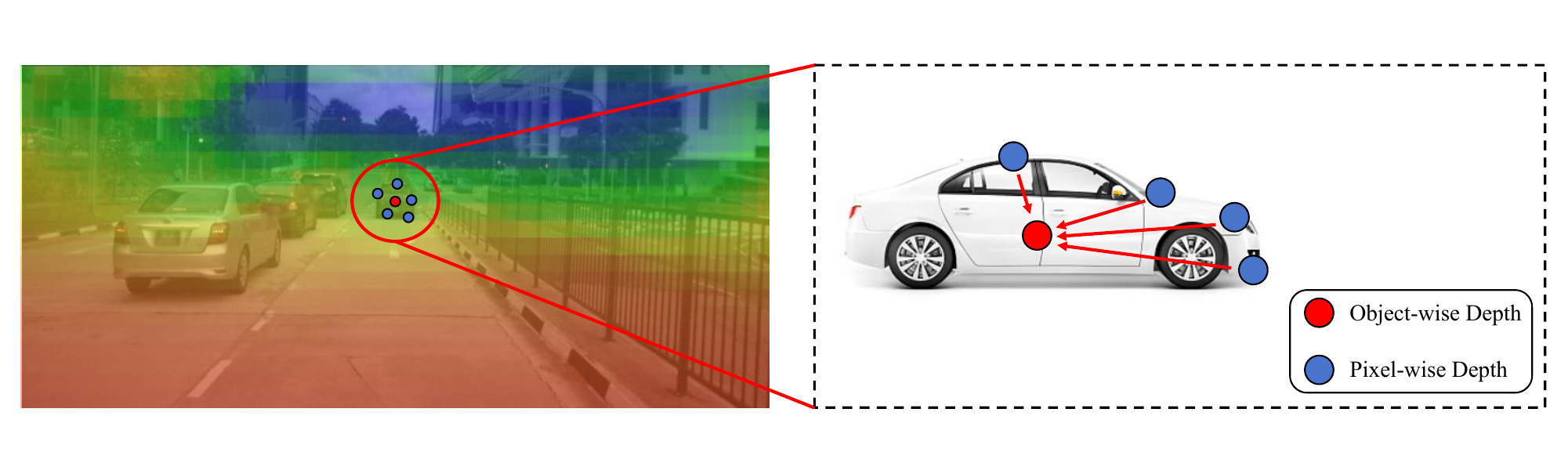}
\caption{The illustration of object-wise depth prediction. The blue points represent the pixel-wise depth, which is usually distributed on the surface of objects and supervised by projected LiDAR points. The red points represent the object-wise depth, which is the 3D center of the object and is supervised by the accurate center of projected 3D ground truth bounding boxes annotated by humans.}
\label{fig_intro_depth}
\vspace{-10pt}
\end{figure}

3D object detection involves the localization and recognition of 3D objects in the real world, which is a fundamental task in 3D perception and is widely applied in autonomous driving and robotics. In recent years, 3D object detection~\cite{huang2021bevdet,li2022bevformer,liu2022petr,li2023bevdepth,li2023bevstereo,shu20233dppe,wang2023exploring} based on camera images has attracted increasing attention due to its lower price than LiDAR sensors. Besides, previous studies proposed many architectures for multi-view 3D object detection and achieved promising performance compared with LiDAR-based 3D object detectors.  

Depth is important information for accurate 3D object detection based on camera images. Recent studies~\cite{li2023bevdepth,shu20233dppe,li2023dfa3d} have explored how to utilize depth information to improve detection performance. In more detail, they usually adopt pixel-wise depth supervision from the projected LiDAR points on the multi-view images to make the network aware of the depth. 
However, we find that the pixel-wise depth supervision obtained from the projected LiDAR points is mainly distributed on the surface of objects, as shown in Figure~\ref{fig_intro_depth}, which is not so friendly to existing DETR-based 3D object detectors. 
Besides, for some distant objects, the fine-grained depth estimation of the whole object is more challenging (\textit{e.g.}, ambiguous for distant objects) than only predicting the depth of 3D centers. 

In contrast, we observe that the object-wise depth representation can alleviate the aforementioned problems, as shown in Figure~\ref{fig_intro_depth}. On the one hand, it is more suitable for DETR-based 3D object detectors since object queries are usually defined as the center of objects. On the other hand, the object-wise depth is more easily estimated accurately, especially for distant objects.
However, there is a new question: \textit{how to effectively introduce the depth information into multi-view 3D object detectors?}
Position embedding is an effective operation to encode the geometric information into multi-view image features, which have been mentioned in existing multi-view 3D object detectors~\cite{liu2022petr,liu2023petrv2,wang2023exploring,shu20233dppe}. Among them, there are two representative position embedding, including ray-aware manner~\cite{wang2023exploring} and point-aware manner~\cite{shu20233dppe}. For ray-aware position embedding, it roughly encodes the depth candidates generated in the camera frustum, leading to an uncertain depth. For point-aware position embedding, it only encodes the predicted pixel-wise depth, which ignores the object-wise depth, leading to sub-optimal performance. 

Towards this goal, in this paper, we propose a new multi-view 3D detector named {\ourMethod}, which mainly introduces the object-wise depth representation to multi-view 3D object detection by our proposed object-wise position embedding. 
Specifically, {\ourMethod} consists of three components: pixel-wise depth encoder (PDE), object-wise depth encoder (ODE), and object-wise position embedding (OPE). PDE first predicts the pixel-wise depth map by aggregating multi-view image features to possess the depth-aware ability of the whole scene. This provides the prior information for the following object-wise depth estimation. Then, the ODE effectively combines the pixel-wise depth representation and temporal information so as to reason the object-wise depth accurately. 

Next, to take full advantage of object-wise depth to enhance detection performance, we encode the object-wise depth into the transformer decoder through our object-wise position embedding, leading to 3D object-aware features.
Besides, we design a depth-aware focal loss (DFL) to encourage the network to pay more attention to the 3D object center.

Overall, our contributions are summarized as follows:
\begin{itemize}
\item We propose a new multi-view 3D object detector named {\ourMethod}, which utilizes the 3D object-wise depth representation to achieve better detection performance. 
\item We introduce the object-wise position embedding to effectively inject object-wise depth information into the transformer decoder, leading to 3D object-aware features. 
\item The proposed {\ourMethod} outperforms previous state-of-the-art methods on the nuScenes~\cite{caesar2020nuscenes} dataset.
\end{itemize}
\section{Related Work}
\label{sec:related_work}

\noindent\textbf{2D Object Detection with DETR.}
DETR~\cite{carion2020end} is the pioneering work that achieves end-to-end object detection by transformer~\cite{vaswani2017attention}. DETR takes learnable object queries as the query and image features as the key and value for the transformer. To achieve end-to-end detection, DETR utilizes Hungarian Matching for ground-truth label assignment. Many works~\cite{zhu2020deformable, meng2021conditional,gao2021fast, wang2022anchor,liu2022dab,li2022dn, zhang2022dino} follow the successful architecture of DETR and further improve the performance. For example, Deformable DETR~\cite{zhu2020deformable} proposed deformable attention, which greatly alleviates the problem of slow convergence of DETR. DINO~\cite{zhang2022dino} utilized denoising training to reduce the learning difficulty of Hungarian Matching.

\noindent\textbf{LiDAR-based 3D Object Detection.}
According to the point cloud processing method, LiDAR-based 3D object detectors can be divided into point-based and voxel-based. Point-based 3D object detectors~\cite{shi2019pointrcnn,qi2019deep,yang2019std,yang20203dssd,he2020structure,he2022voxel,zhang2022not} usually randomly sample point clouds and directly consume sampled point clouds for extracting 3D features by a PointNet-like~\cite{qi2017pointnet} backbone. For voxel-based 3D object detectors~\cite{zhou2018voxelnet,yan2018second,lang2019pointpillars,shi2020pv,deng2021voxel,shi2021pv}, these methods first quantify point clouds into regular grids and then utilize a 3D sparse convolution backbone to extract 3D features.

\noindent\textbf{Multi-view 3D Object Detection.}
Multi-view 3D object detectors consume multi-view images from surrounding camera sensors and detect 3D objects in the 3D space. According to the view transformation paradigm, multi-view 3D object detectors can be divided into LSS-based and transformer-based. For LSS-based multi-view 3D object detectors~\cite{huang2021bevdet,li2023bevdepth,li2023bevstereo}, these methods utilize Lift-Splat-Shot (LSS) operation~\cite{philion2020lift} to convert multi-view image features to bird's-eye view (BEV) feature and BEV-based 3D detection head to detect objects on the BEV feature. For transformer-based methods~\cite{li2022bevformer,liu2022petr,wang2023exploring,xiong2023cape}, these methods utilize the transformer to convert multi-view image features to 3D features. For example, BEVFormer~\cite{li2022bevformer} utilizes Deformable attention~\cite{zhu2020deformable} to aggregate multi-view image features to BEV feature. PETR~\cite{liu2022petr} utilizes 3D position embedding to project multi-view image features to 3D space. Depth is important information to achieve accurate 3D object detection. Therefore, many works explore the utilization of depth to improve detection performance. BEVDepth~\cite{li2023bevdepth} and BEVStereo~\cite{li2023bevstereo} utilize projected LiDAR points to supervise depth prediction, which is used to enhance the LSS and improve performance. DFA3D~\cite{li2023dfa3d} proposes the 3D deformable attention based on the predicted depth map to achieve better view transformation. 3DPPE~\cite{shu20233dppe} proposes point-aware position embedding by 3D points generated by the predicted depth map and pixel coordinates to 3D space. 

However, previous methods ignored the problems of the predicted depth map supervised by projected LiDAR points. In this paper, we propose a new multi-view 3D object detector named {\ourMethod}, which utilizes the object-wise position embedding to effectively inject object-wise depth information into the network through our proposed object-wise position embedding and enhance the interaction with object queries in the transformer decoder. Benefiting the object-wise depth information, our {\ourMethod} achieves the state-of-the-art performance for 3D object detection on the nuScenes~\cite{caesar2020nuscenes} dataset.

\begin{figure}[t!]
\centering
    \includegraphics[width=0.99\linewidth]{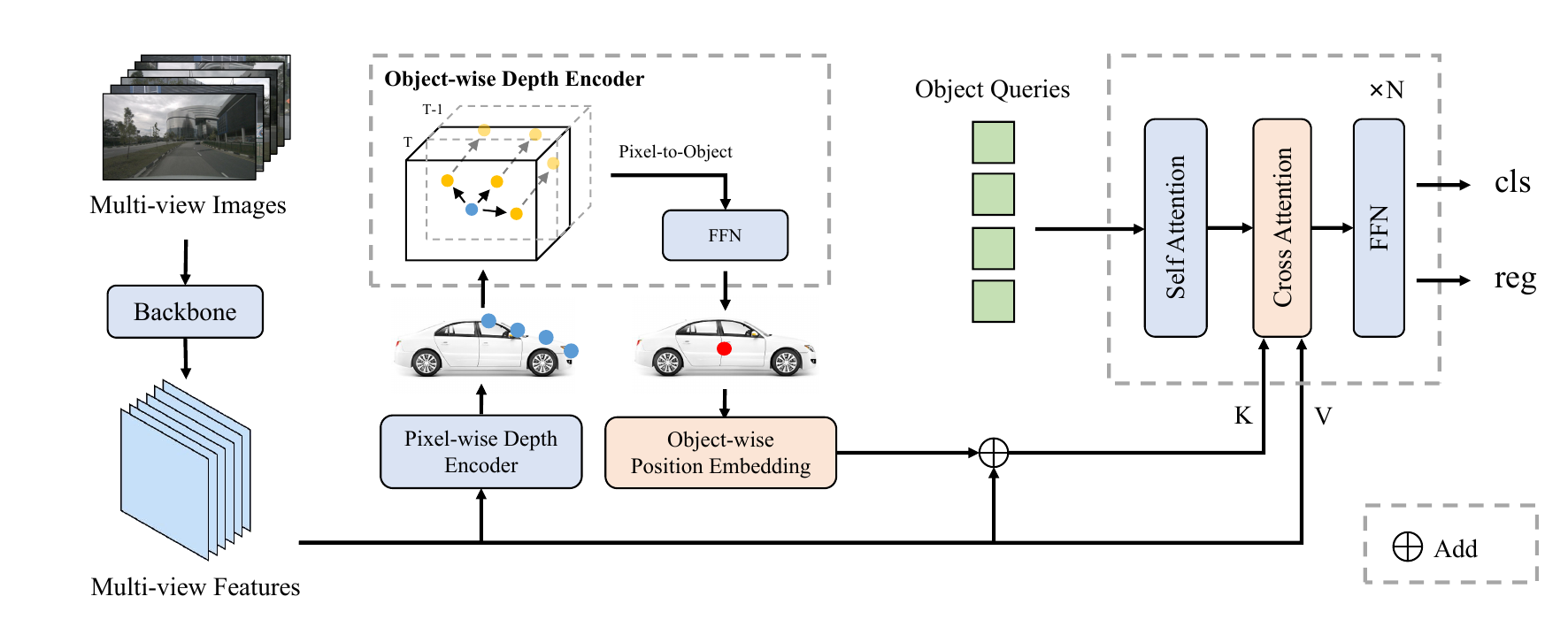}
\caption{The overall architecture of the proposed {\ourMethod}, which consists of the pixel-wise depth encoder (PDE), the object-wise depth encoder (ODE), and object-wise position embedding (OPE). Specifically, the PDE first utilizes a DepthNet to predict the pixel-wise depth map supervised by projected LiDAR points. Then, the ODE predicts the object-wise depth, supervised by the center of projected 3D bounding boxes, based on the predicted pixel-wise depth map. Finally, {\ourMethod} utilizes the object-wise position embedding based on predicted object-wise depth and corresponding 2D object centers to convert the multi-view image features to object-wise 3D features for interaction with object queries and generate final detection results.}
\label{fig_arch}
\end{figure}

\section{Method}
\label{sec:method}

\subsection{Overall Architecture}
Many existing works~\cite{li2023bevdepth,shu20233dppe,li2023dfa3d} also utilize the projected LiDAR points to supervise the depth prediction. However, the depth supervision obtained from the LiDAR points is usually distributed on the surface of the object, which is not so friendly to existing DETR-based 3D object detectors, and the fine-grained depth estimation of the whole object is more challenging, especially for some distant objects. Therefore, we propose a new multi-view 3D object detector named {\ourMethod}, whose overall architecture is presented in Figure~\ref{fig_arch}. Specifically, given $N$ multi-view images,  {\ourMethod} first utilizes a 2D backbone to extract $N$ multi-view features. Then, for $N$ views, {\ourMethod} follow DepthNet~\cite{shu20233dppe} to predict the pixel-wise depth map supervised by projected LiDAR points. Next, the object-wise depth encoder (ODE) combines the pixel-wise depth representation and temporal information to reason the object-wise depth accurately. Finally, {\ourMethod} encodes the object-wise depth into the transformer decoder through our object-wise position embedding (OPE) to predict final detection results. 

\subsection{Pixel-wise Depth Encoder}
In order to obtain a more accurate object-wise depth prediction, {\ourMethod} first utilizes PDE to predict the pixel-wise depth map, which serves as the depth prior of subsequent object-wise depth prediction and is supervised by projected LiDAR points. Given multi-view features $\mathbf{F} = \{\mathbf{F}_i \in \mathbb{R}^{C \times H \times W}, i=1,2,3,...,N\}$ generated from $N$ multi-view images, where $C$, $H$, and $W$ are the channel dimension, height, and width of image features. PDE first encodes camera intrinsic $\mathbf{K} \in \mathbb{R}^{4\times 4}$ by an MLP for modulating $\mathbf{F}$. Then, it utilizes a DepthNet~\cite{shu20233dppe}, which consists of several residual blocks and deformable convolutions, to predict the pixel-wise depth map. Besides, we follow the fusion depth~\cite{shu20233dppe}, which combines regression depth and probabilistic depth to generate the final pixel-wise depth map $\mathbf{D}_{i} \in \mathbb{R}^{H \times W \times 1}$ for the following object-wise depth encoder. 

\subsection{Object-wise Depth Encoder}
\begin{figure}[t!]
\centering
    \includegraphics[width=0.99\linewidth]{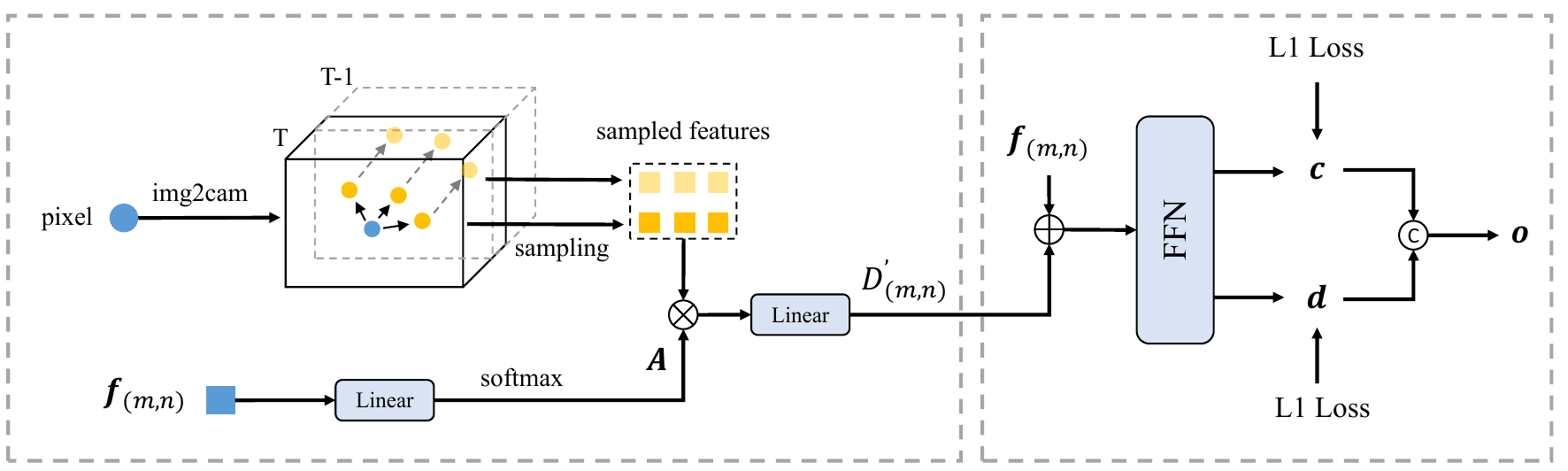}
\caption{The overall architecture of the ODE. ODE first converts image pixels from the pixel coordinate to the camera coordinate and aggregates current and historical features to generate depth embedding for object-wise depth prediction by streaming temporal fusion strategy. Finally, ODE utilizes an FFN to predict the object-wise depth $d$ and corresponding object center $c$ based on the depth embedding.}
\label{fig_ode}
\end{figure}

After obtaining the pixel-wise depth map, ODE performs an attention-based temporal depth aggregation to predict object-wise depth. The object-wise depth is supervised by the projected 3D ground truth bounding boxes.
Given the predicted pixel-wise depth $\mathbf{D}_{(m,n)}$, where $(m,n)$ represents the pixel in the $m^{th}$ row and $n^{th}$ column on the $i^{th}$ image. As shown in Figure~\ref{fig_ode}, ODE converts image pixel coordinates $(u,v)^{\mathrm{T}}$ to the camera coordinate based on the predicted pixel-wise depth $\mathbf{D}_{(m,n)}$. The transformation is formulated as:
\begin{equation}
\mathbf{p}^{'}_{(m,n)}=(u \times \mathbf{D}_{(m,n)}, v \times \mathbf{D}_{(m,n)}, \mathbf{D}_{(m,n)}, 1)^{\mathrm{T}},
\end{equation}
\begin{equation}
\mathbf{p}_{(m,n)}=\mathbf{K}^{-1} \mathbf{p}^{'}_{(m,n)},
\end{equation}
where $\mathbf{p}_{(m,n)}$ is the projected point on the camera coordinate, which will be used as the reference point.
After generating a series of reference points $\mathbf{P}=\{\mathbf{p}_{(m,n)},m=1,2,3,...,H \  \mathrm{and} \ n=1,2,3,...,W\}$. ODE first generates pixel-wise depth embedding and then combines this embedding and image features to predict the object-wise depth. Specifically, given the reference point $\mathbf{p}_{(m,n)}$ and the corresponding image feature $\mathbf{f}_{(m,n)}$, ODE utilizes a linear function to predict $k$ 3D offsets, which are added to $\mathbf{p}_{(m,n)}$ to obtain 3D sampling points. Then, we project 3D sampling points to the pixel coordinate of the current and previous frames. These projected points $\mathbf{p}^{*}_{(m,n)}$ are used to sample corresponding features. Meanwhile, the attention weight $\mathbf{A}$ is predicted by feeding $\mathbf{f}_{(m,n)}$ into a linear and softmax function. Finally, the sampled features $\mathbf{f}_s$ is multiplied with $\mathbf{A}$ for generating pixel-wise depth embedding $\mathbf{E}_{(m,n)}$. We can formulate the above process as follows:
\begin{equation}
\mathbf{E}_{(m,n)} = \phi(\sum\limits_{j=1}^{k} \mathbf{A}_{j} \cdot \mathrm{Concat}(\mathbf{F}_{i}(\mathbf{p}^{*}_{(m,n)}), \mathbf{F}^{'}_{i}(\mathbf{p}^{*}_{(m,n)}))),
\end{equation}
where $\phi$ is a linear function. $\mathrm{Concat}$ denotes the concatenation operation. $\mathbf{F}_{i}(*)$ and $\mathbf{F}^{'}_{i}(*)$ denote sampling the corresponding features of the coordinates $*$ on the current image features $\mathbf{F}$ and previous image features $\mathbf{F}^{'}$ by a bilinear interpolation operation, respectively. 

Finally, the pixel-wise depth embedding $\mathbf{E}_{(m,n)}$ and image features $\mathbf{F}_{i}$ are fed into a FFN to predict the object-wise depth $\mathbf{d} \in \mathbb{R}^{(H \times W) \times 1}$ and corresponding object center $\mathbf{c} \in \mathbb{R}^{(H \times W) \times 2}$ on the image.

\subsection{Object-wise Position Embedding}
To utilize the generated object-wise depth, we present OPE to encode the object-wise depth into the transformer decoder to enhance 3D-aware features and improve the performance of multi-view 3D object detection. Next, we will discuss the differences between the different position embedding methods and introduce the details of OPE.

\begin{figure}[t!]
\centering
    \includegraphics[width=0.99\linewidth]{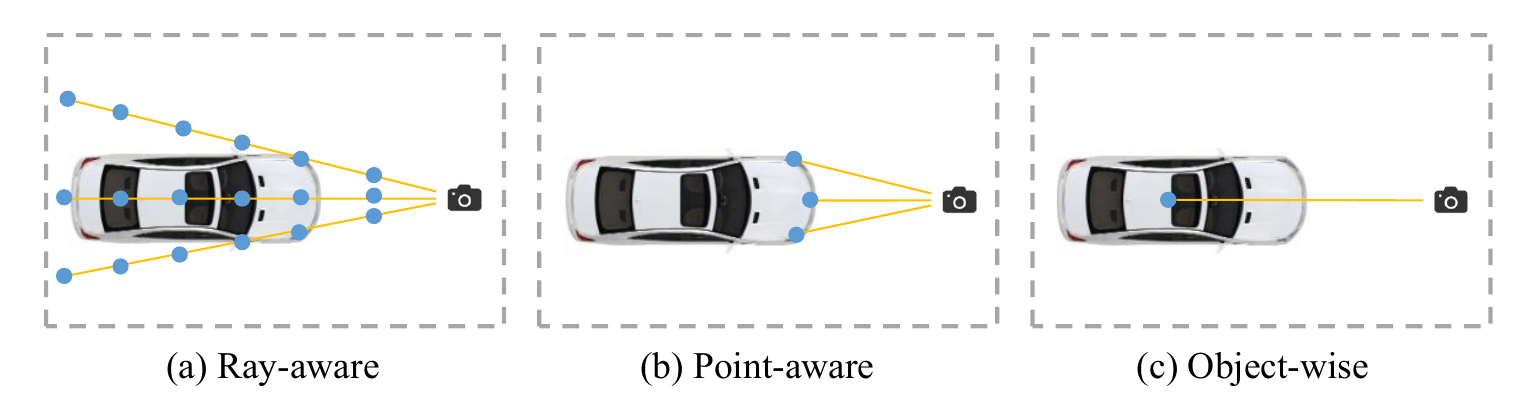}
\caption{Comparison of the ray-aware position embedding (a), point-aware position embedding (b), and the object-wise position embedding (c). Compared with other methods, OPE utilizes the 3D object center to generate the position embedding, which can achieve better 3D representation.}
\label{fig_cmp_pe}
\end{figure}

\noindent\textbf{Discussion of Different Position Embedding.}
As shown in Figure~\ref{fig_cmp_pe}, we compare the ray-aware position embedding of StreamPETR~\cite{wang2023exploring}, the point-aware position embedding of 3DPPE~\cite{shu20233dppe}, and our object-wise position embedding (OPE). The ray-aware position embedding (a) first generates a 3D mesh grid in the camera frustum space based on depth candidates and then converts these points to the LiDAR coordinate as the position embedding. The point-aware position embedding (b) first utilizes a DepthNet~\cite{shu20233dppe} to predict the pixel-wise depth map and then converts these 3D points generated by the predicted depth map to the LiDAR coordinate as the position embedding. Although these methods achieve satisfactory performance, there are still ignored problems. For ray-aware position embedding, the uncertainty depth estimation without depth supervision makes it difficult to generate accurate 3D-aware features. For point-aware position embedding, although it adopts projected LiDAR points to supervise the pixel-wise depth prediction and encodes 3D points for position embedding to improve performance, it ignores the importance of object-wise depth for DETR-based 3D object detectors, leading to sub-optimal performance. Therefore, we propose object-wise position embedding (OPE) to alleviate the above problems. Compared with other methods, the OPE can generate more accurate 3D-aware features based on object-wise depth representation, resulting in better detection performance.

\noindent\textbf{Details of OPE.} Given the $j^{th}$ object-wise depth $d_j$ and corresponding object center $(x,y)$ predicted by ODE on the $i^{th}$ image, OPE generates object position $\mathbf{o}_j=(x,y,d_j)$. After generating $\mathbf{o}_j$, OPE converts $\mathbf{o}_j$ in the pixel coordinate to the 3D object center $\mathbf{O}_j$ in the LiDAR coordinate:
\begin{equation}
\mathbf{o}^{'}_j=(x \times d_j, y \times d_j, d_j, 1)^{\mathrm{T}},
\end{equation}
\begin{equation}
\mathbf{O}_j=\mathbf{R}^{-1} \mathbf{K}^{-1} \mathbf{o}^{'}_j,
\end{equation}
where $\mathbf{K}$ and $\mathbf{R}$ denote the camera intrinsic and extrinsic.
Then, OPE normalizes the 3D object center by the perception range and utilizes a 3D position embedding operation to encode the 3D object center in a cosine manner. Finally, OPE adopts a multi-layer perceptron (MLP) to generate the object-wise position embedding:
\begin{equation}
\mathbf{OPE}_j=\mathrm{MLP}((\mathrm{PE}_{3D}(\mathrm{Norm}(\mathbf{O}_j)))),
\end{equation}
where $\mathbf{OPE}_j$, $\mathrm{Norm}$, and $\mathrm{PE}_{3D}$ denote the $j^{th}$ object-wise position embedding, normalization, and 3D position embedding operation respectively. Finally, we add $\mathbf{OPE}_j$ to the $j^{th}$ image feature for feature interaction with learnable object queries in the transformer decoder.

\subsection{Depth-aware Focal Loss}
DFL aims to further encourage {\ourMethod} to pay more attention to the object center. Different from the traditional focal loss~\cite{lin2017focal}, DFL takes the depth score $s$ as the classification label to generate a soft label for focal loss. Specifically, given the predicted 3D object center $\hat{\mathbf{C}}$. the corresponding ground truth $\mathbf{C}$, the binary target class label $\mathbf{t}$, and the predicted classification probability $\hat{\mathbf{p}}$, then the DFL is formulated as:
\begin{equation}
\mathbf{\mathcal{L}}_{DFL}=-\bm{\alpha}^{'} \cdot \lvert \mathbf{t} \cdot \mathbf{s} - \hat{\mathbf{p}}\rvert^{\gamma} \cdot \mathrm{log}(\lvert 1 - \mathbf{t} - \hat{\mathbf{p}}\rvert),
\end{equation}
where $\mathbf{s}=e^{-\mathrm{L2}(\hat{\mathbf{C}} - \mathbf{C})}$, $\bm{\alpha}^{'}=\alpha \cdot \mathbf{t} \cdot \mathbf{s} + (1-\alpha) \cdot (1-\mathbf{t} \cdot \mathbf{s})$. $\mathrm{L2}$ is the euclidean distance. $\alpha$ and $\gamma$ are hyper-parameters. 

\noindent\textbf{Total Loss.}
Given the depth-aware focal loss $\mathbf{\mathcal{L}}_{DFL}$, 3D bounding box regression loss $\mathbf{\mathcal{L}}_{reg}$, the pixel-wise depth prediction loss $\mathbf{\mathcal{L}}_{PDE}$, and the object-wise depth prediction loss $\mathbf{\mathcal{L}}_{ODE}$, we adopt the Hungarian Matching to achieve label assignment and the total loss can be calculated as:
\begin{equation}
\mathbf{\mathcal{L}}=\lambda_{1} \mathbf{\mathcal{L}}_{PDE} + \lambda_{2} \mathbf{\mathcal{L}}_{ODE} + \lambda_{3} \mathbf{\mathcal{L}}_{DFL} + \lambda_{4} \mathbf{\mathcal{L}}_{reg},
\end{equation}
where $\lambda_{1}$, $\lambda_{2}$, $\lambda_{3}$, and $\lambda_{4}$ are hyper-parameter to balance different losses.

\section{Experiments}
\label{sec:exp}

\subsection{Datasets and Metrics}
We validate our method on the nuScenes~\cite{caesar2020nuscenes} benchmark. nuScenes is the widely used autonomous driving dataset for multi-view 3D object detection. The nuScenes dataset consists of 1000 scenes, which are divided into three parts: 750 for training, 150 for validation, and 150 for testing, where each scene is roughly 20s long and annotated at 2Hz. The nuScenes dataset provides point clouds collected by 32-beam LiDAR and 6 multi-view images collected by 6 surrounding cameras. Besides, the evaluation of the nuScnes dataset adopts the Mean Average Precision~(mAP) and nuScenes detection score~(NDS) to evaluate the performance of 3D detectors for 10 foreground classes. Consistent with previous methods, we report NDS and mAP, along with mean Average Translation Error (mATE), mean Average Scale Error (mASE), mean Average Orientation Error(mAOE), mean Average Velocity Error(mAVE), and mean Average
Attribute Error(mAAE).

\subsection{Implementation Details}
We follow StreamPETR~\cite{wang2023exploring} to conduct experiments with ResNet50~\cite{he2016deep}, ResNet101, and V2-99~\cite{wang2021fcos3d} backbones on the nuScenes~\cite{caesar2020nuscenes} dataset without any test-time augmentation or future information. {\ourMethod} is trained with the nuImages~\cite{caesar2020nuscenes} pre-trained model for the nuScenes \texttt{val} set and with the DD3D~\cite{park2021pseudo} pre-trained model for the nuScenes \texttt{test} set. For the pixel-wise depth encoder, we take the $8\times$ downsampled depth map generated by projected LiDAR points as the supervision. For the object-wise depth encoder, we set the number of prediction offsets $k$ to 13. For the depth-aware focal loss, we set $\alpha$ to 0.25 and $\gamma$ to 2.0. For the balancing factors of different losses, we set $\lambda_1$, $\lambda_2$, $\lambda_3$, and $\lambda_4$ to 1.0, 5.0, 2.0 and 0.25. 

For training {\ourMethod}, we use the same data augmentation methods as PETR~\cite{liu2022petr} and AdamW~\cite{loshchilov2017decoupled} optimizer with a batch size of 16 on 8 NVIDIA Tesla V100 GPUs without CBGS~\cite{zhu2019class} strategy. For comparison with other methods on the nuScenes \texttt{val} set, we use the streaming video training method~\cite{wang2023exploring} to train {\ourMethod} for 90 epochs with a starting learning rate of $4e^{-4}$ and cosine annealing policy. It is worth noting that compared with the sliding window training method, the streaming video training method needs more epochs for convergence but still saves much training time. For comparison with other methods on the nuScenes \texttt{test} set, we train 60 epochs to prevent over-fitting.

\subsection{Main Results}

\begin{table*}[t!]
\centering
\caption{
Comparison of other methods on the nuScenes \texttt{val} set. ${}^{\dag}$ The backbone benefits from perspective pretraining.
}
\resizebox{1.0\linewidth}{!}{
\begin{tabular}{l|c|c|cc|ccccc}
\toprule
Method & Backbone & Input Size & NDS$\uparrow$ & mAP$\uparrow$ & mATE$\downarrow$  & mASE$\downarrow$ & mAOE$\downarrow$ & mAVE$\downarrow$ & mAAE$\downarrow$ \\
\midrule
BevDet4D~\cite{huang2022bevdet4d} & ResNet50 & $256 \times 704$ & 45.7 & 32.2 & 0.703 & 0.278 & 0.495 & 0.354 & 0.206 \\
PETRv2~\cite{liu2023petrv2} & ResNet50 & $256 \times 704$ & 45.6 & 34.9 & 0.700 & 0.275 & 0.580 & 0.437 & 0.187 \\
BEVDepth~\cite{li2023bevdepth} & ResNet50 & $256 \times 704$ & 47.5 & 35.1 & 0.639 & 0.267 & 0.479 & 0.428 & 0.198 \\
BEVStereo~\cite{li2023bevstereo} & ResNet50 & $256 \times 704$ & 50.0 & 37.2 & 0.598 & 0.270 & 0.438 & 0.367 & 0.190 \\
BEVFormerv2$^{\dag}$~\cite{yang2023bevformer} & ResNet50 & - & 52.9 & 42.3 & 0.618 & 0.273 & 0.413 & 0.333 & 0.188 \\
SOLOFusion~\cite{park2022time} & ResNet50 & $256 \times 704$ & 53.4 & 42.7 & 0.567 & 0.274 & 0.511 & 0.252 & 0.181 \\
Sparse4Dv2~\cite{lin2023sparse4d} & ResNet50 & $256 \times 704$ & 53.8 & 43.9 & 0.598 & 0.270 & 0.475 & 0.282 & 0.179 \\
StreamPETR$^{\dag}$~\cite{wang2023exploring} & ResNet50 & $256 \times 704$ & 55.0 & 45.0 & 0.613 & 0.267 & 0.413 & 0.265 & 0.196 \\
SparseBEV$^{\dag}$~\cite{liu2023sparsebev} & ResNet50 & $256 \times 704$ & 55.8 & 44.8 & 0.581 & 0.271 & 0.373 & 0.247 & 0.190 \\
\rowcolor{gray!20}
{\ourMethod}$^{\dag}$ & ResNet50 & $256 \times 704$ & \textbf{56.4} & \textbf{46.5} & 0.573 & 0.275 & 0.413 & 0.235 & 0.193 \\
\midrule
3DPPE~\cite{shu20233dppe} & ResNet101 & $512 \times 1408$ & 45.8 & 39.1 & 0.674 & 0.282 & 0.395 & 0.830 & 0.191 \\
BEVDepth~\cite{li2023bevdepth} & ResNet101 & $512 \times 1408$ & 53.5 & 41.2 & 0.565 & 0.266 & 0.358 & 0.331 & 0.190 \\
SOLOFusion~\cite{park2022time} & ResNet101 & $512 \times 1408$ & 58.2 & 48.3 & 0.503 & 0.264 & 0.381 & 0.246 & 0.207 \\
SparseBEV$^{\dag}$~\cite{liu2023sparsebev} & ResNet101 & $512 \times 1408$ & 59.2 & 50.1 & 0.562 & 0.265 & 0.321 & 0.243 & 0.195 \\
StreamPETR$^{\dag}$~\cite{wang2023exploring} & ResNet101 & $512 \times 1408$ & 59.2 & 50.4 & 0.569 & 0.262 & 0.315 & 0.257 & 0.199 \\
Sparse4Dv2$^{\dag}$~\cite{lin2023sparse4d} & ResNet101 & $512 \times 1408$ & 59.4 & 50.5 & 0.548 & 0.268 & 0.348 & 0.239 & 0.184 \\
Far3D$^{\dag}$~\cite{jiang2023far3d} & ResNet101 & $512 \times 1408$ & 59.4 & 51.0 & 0.551 & 0.258 & 0.372 & 0.238 & 0.195 \\
\rowcolor{gray!20}
{\ourMethod}$^{\dag}$ & ResNet101 & $512 \times 1408$ & \textbf{60.6} & \textbf{51.6} & 0.528 & 0.266 & 0.312 & 0.222 & 0.190 \\
\bottomrule
\end{tabular}
}
\label{tab:val}
\end{table*}

\begin{table*}[t!]
\centering
\caption{
Comparison of other methods on nuScenes \texttt{test} set. These results are reported without test-time augmentation, model ensembling, and any future information.
}
\resizebox{1.0\linewidth}{!}{
\begin{tabular}{l|c|c|cc|ccccc}
\toprule
Method & Backbone & Input Size & NDS$\uparrow$ & mAP$\uparrow$ & mATE$\downarrow$  & mASE$\downarrow$ & mAOE$\downarrow$ & mAVE$\downarrow$ & mAAE$\downarrow$\\
\midrule
BEVDepth~\cite{li2023bevdepth} & V2-99 & $640\times1600$ & 60.0 & 50.3 & 0.445 & 0.245 & 0.378 & 0.320 & 0.126 \\
BEVStereo~\cite{li2023bevstereo} & V2-99 & $640\times1600$ & 61.0 & 52.5 & 0.431 & 0.246 & 0.358 & 0.357 & 0.138 \\
CAPE-T~\cite{xiong2023cape} & V2-99 & $640\times1600$ & 61.0 & 52.5 & 0.503 & 0.242 & 0.361 & 0.306 & 0.114 \\
FB-BEV~\cite{liu2023sparsebev} & V2-99 & $640\times1600$ & 62.4 & 53.7 & 0.439 & 0.250 & 0.358 & 0.270 & 0.128 \\
HoP~\cite{zong2023temporal} & V2-99 & $640\times1600$ & 61.2 & 52.8 & 0.491 & 0.242 & 0.332 & 0.343 & 0.109 \\
StreamPETR~\cite{wang2023exploring} & V2-99 & $640\times1600$  & 63.6 & 55.0 & 0.479 & 0.239 & 0.317 & 0.241 & 0.119 \\
SparseBEV~\cite{liu2023sparsebev} & V2-99 & $640\times1600$  & 63.6 & 55.6 & 0.485 & 0.244 & 0.332 & 0.246 & 0.117 \\
Sparse4Dv2~\cite{lin2023sparse4d} & V2-99 & $640\times1600$ & 63.8 & 55.6 & 0.462 & 0.238 & 0.328 & 0.264 & 0.115 \\
\rowcolor{gray!20}
{\ourMethod} & V2-99 & $640\times1600$ & \textbf{64.4} & \textbf{56.7} & 0.456 & 0.244 & 0.325 & 0.240 & 0.129 \\
\bottomrule
\end{tabular}
}
\label{tab:test}
\end{table*}

We compare the proposed {\ourMethod} with previous state-of-the-art multi-view 3D object detectors on the nuScenes \texttt{val} and \texttt{test} sets. As shown in Table~\ref{tab:val}, for nuScenes \texttt{val} set, {\ourMethod} achieves 56.4\% NDS and 46.5\% mAP performance with an image size of $256\times 704$ and ResNet50 backbone pre-trained on nuImages, which outperforms the state-of-the-art method (SparseBEV) by 0.6\% NDS and 0.7\% mAP, and outperforms our baseline (StreamPETR) by 1.4\% NDS and 1.5\% mAP. When adopting an image size of $512\times 1408$ and ResNet101 backbone pre-trained on nuImage, {\ourMethod} achieves 60.6\% NDS and 51.6\% mAP performance, which yields a new state-of-the-art result on the nuScenes \texttt{val} set.  {\ourMethod} exceeds the state-of-the-art method (Far3D) by 1.2\% NDS and 0.6\% mAP. Finally, our {\ourMethod} has an obvious performance improvement over our baseline with NDS of 1.4\%  and mAP of 1.2\%. 

Furthermore, we provide the detection results on nuScenes \texttt{test} set, shown in Table~\ref{tab:test}. We observe that {\ourMethod} achieves 64.4\% NDS and 56.7\% mAP under the image size of $640\times 1600$ based on V2-99~\cite{wang2021fcos3d} backbone pre-trained on DD3D, which yields a new state-of-the-art result on the nuScenes \texttt{test} set.
These experiments clearly demonstrate the effectiveness of our method.

\subsection{Ablation Study}
In this section, we conduct ablation studies to investigate the effectiveness of {\ourMethod} on the nuScenes \texttt{val} set. If not specified, we adopt the V2-99 backbone pre-trained on DD3D with an image size of $320\times 800$. Here, we train {\ourMethod} for 24 epochs in ablation studies.  

\begin{table*}[t]
\centering
\caption{
Ablation studies for each component in {\ourMethod} on the nuScenes \texttt{val} set. The PDE, ODE, OPE, and DFL represent the pixel-wise depth encoder, object-wise depth encoder, object-wise position embedding, and depth-aware focal loss, respectively.
}
\begin{tabular}{c|cccc|cc|ccccc}
\toprule
\# & PDE & ODE & OPE & DFL & NDS$\uparrow$ & mAP$\uparrow$ & mATE$\downarrow$  & mASE$\downarrow$ & mAOE$\downarrow$ & mAVE$\downarrow$ & mAAE$\downarrow$ \\
\midrule
\uppercase\expandafter{\romannumeral1} & & & & & 59.4 & 50.3 & 0.575 & 0.258 & 0.300 & 0.243 & 0.196 \\
\uppercase\expandafter{\romannumeral2} & \checkmark & & & & 59.4 & 50.5 & 0.564 & 0.257 & 0.320 & 0.252 & 0.190 \\
\uppercase\expandafter{\romannumeral3} & \checkmark & \checkmark & & & 59.7 & 50.6 & 0.568 & 0.257 & 0.305 & 0.245 & \textbf{0.187} \\
\uppercase\expandafter{\romannumeral4} & \checkmark & \checkmark & \checkmark & & 60.8 & \textbf{52.4} & 0.553 & 0.258 & 0.291 & 0.242 & 0.197\\
\uppercase\expandafter{\romannumeral5} & \checkmark & \checkmark & \checkmark & \checkmark & \textbf{61.3} & 52.1 & \textbf{0.525} & \textbf{0.256} & \textbf{0.281} & \textbf{0.216} & 0.199 \\
\bottomrule
\end{tabular}
\label{tab:ablation}
\end{table*}

\noindent\textbf{Effectiveness of Each Component.}
As shown in Table~\ref{tab:ablation}, we adopt the StreamPETR~\cite{wang2023exploring} as our baseline (\uppercase\expandafter{\romannumeral1}) and conduct ablation studies on each component in {\ourMethod}. Compared with the baseline, the PDE (\uppercase\expandafter{\romannumeral2}) only brings 0.2\% mAP performance improvements, indicating that pixel-wise depth supervision can not significantly boost detection performance. Compared with (\uppercase\expandafter{\romannumeral2}), the ODE (\uppercase\expandafter{\romannumeral3}) further brings 0.3\% NDS and 0.1\% mAP performance improvements. Compared with (\uppercase\expandafter{\romannumeral3}), when adopting the OPE (\uppercase\expandafter{\romannumeral4}) in the transformer decoder, the performance is further improved by 1.1\% NDS and 1.8\% mAP, demonstrating the effectiveness of encoding object-wise depth information into the network by our proposed OPE. Compared with (\uppercase\expandafter{\romannumeral4}), the DFL (\uppercase\expandafter{\romannumeral5}) can further bring 0.5\% NDS performance improvements by paying more attention to the 3D object center information. 
Finally, combining all components, {\ourMethod} achieves significant performance improvements (1.9\% NDS and 1.8\% mAP) over the baseline. These experiments demonstrate the effectiveness of the proposed component. 

\begin{table*}[t]
\centering
\caption{
Ablation studies of the object-wise depth encoder on the nuScenes \texttt{val} set. The temporal denotes the utilization of temporal information in ODE.
}
\setlength{\tabcolsep}{10pt}
\resizebox{1.0\linewidth}{!}{
\begin{tabular}{c|cc|ccccc}
  \toprule
    temporal & NDS$\uparrow$ & mAP$\uparrow$ & mATE$\downarrow$  & mASE$\downarrow$ & mAOE$\downarrow$ & mAVE$\downarrow$ & mAAE$\downarrow$\\
    \midrule
      & 61.0 & 51.7 & 0.530 & \textbf{0.256} & \textbf{0.280} & 0.218 & 0.201 \\
     \checkmark & \textbf{61.3} & \textbf{52.1} & \textbf{0.525} & \textbf{0.256} & 0.281 & \textbf{0.216} & \textbf{0.199} \\
    \bottomrule
\end{tabular}
}
\label{tab:ode}
\end{table*}

\noindent\textbf{Ablations of the ODE.}
The object-wise depth encoder combines the temporal information to predict the object-wise depth accurately based on the pixel-wise depth map. Here, we explore the effect of the temporal information. As shown in Table~\ref{tab:ode}, when ODE disables the temporal information, we find there is a 0.3\% NDS and 0.4\% mAP performance drop. We think that the temporal information is helpful to improve the accuracy of the object-wise depth prediction. This experiment effectively illustrates the necessity of our ODE by utilizing temporal information.

\begin{table*}[t]
\centering
\caption{
Comparison of other position embedding methods on the nuScenes \texttt{val} set. The Ray-aware PE, Point-aware PE, and OPE denote the ray-aware position embedding, point-aware position embedding, and object-wise position embedding, respectively.
}
\resizebox{1.0\linewidth}{!}{
\begin{tabular}{c|cc|ccccc}
\toprule
Method & NDS$\uparrow$ & mAP$\uparrow$ & mATE$\downarrow$  & mASE$\downarrow$ & mAOE$\downarrow$ & mAVE$\downarrow$ & mAAE$\downarrow$ \\
\midrule
Ray-aware PE~\cite{wang2023exploring} & 59.4 & 50.3 & 0.575 & \textbf{0.258} & 0.300 & 0.243 & \textbf{0.196} \\
Point-aware PE~\cite{shu20233dppe} & 60.0 (\textcolor{up}{+0.6}) & 51.6 (\textcolor{up}{+1.3}) & 0.568 & \textbf{0.258} & 0.306 & 0.245 & 0.203 \\
OPE & \textbf{60.8} (\textbf{\textcolor{up}{+1.4}}) & \textbf{52.4} (\textbf{\textcolor{up}{+2.1}}) & \textbf{0.553} & \textbf{0.258} & \textbf{0.291} & \textbf{0.242} & 0.197\\
\bottomrule
\end{tabular}
}
\label{tab:ope}
\end{table*}

\begin{table*}[t]
\centering
\caption{
Comparison of other position embedding methods on the nuScenes \texttt{val} set in different distances. NDS$_{\textgreater0}$, NDS$_{\textgreater20}$, and NDS$_{\textgreater40}$ represent different evaluation metrics under distance thresholds of 0, 20, and 40 meters, respectively.
}

\setlength{\tabcolsep}{20pt}
\resizebox{1.0\linewidth}{!}{
   \begin{tabular}{c|c|c|c}
        \toprule
        Method & NDS$_{\textgreater 0}$$\uparrow$ & NDS$_{\textgreater 20}$$\uparrow$ & NDS$_{\textgreater 40}$$\uparrow$\\
        \midrule
        Ray-aware PE~\cite{wang2023exploring} & 59.4 & 49.5 & 36.8\\
        Point-aware PE~\cite{shu20233dppe} & 60.0 (\textcolor{up}{+0.6}) & 50.4 (\textcolor{up}{+0.9})& 37.9 (\textcolor{up}{+1.1})\\
        OPE & \textbf{60.8} (\textbf{\textcolor{up}{+1.4}}) & \textbf{51.0} (\textbf{\textcolor{up}{+1.5}}) & \textbf{39.1} (\textbf{\textcolor{up}{+2.3}})\\
    \bottomrule
    \end{tabular}
}
\label{tab:dist}
\vspace{-10pt}
\end{table*}

\noindent\textbf{Effectiveness of the OPE.}
To further verify the effectiveness of OPE, we compare different position embedding methods on the nuScenes \texttt{val} set. For StreamPETR~\cite{wang2023exploring}, it generates a 3D mesh grid in the camera frustum space based on depth candidates and converts these points to the LiDAR coordinate as the ray-aware position embedding. For 3DPPE~\cite{shu20233dppe}, it generates 3D points based on the predicted pixel-wise depth map and converts these points to the LiDAR coordinate as the point-aware position embedding. For OPE, it generates object centers based on predicted object-wise depth and converts them to the LiDAR coordinate as the object-wise position embedding. As shown in Table~\ref{tab:ope}, when adopting the ray-aware position embedding (\uppercase\expandafter{\romannumeral1}), the performance is 59.4\% NDS and 50.3\% mAP. The point-aware position embedding (\uppercase\expandafter{\romannumeral2}) can bring 0.6\% NDS and 1.3\% mAP compared with the ray-aware position embedding. After utilizing the proposed object-wise position embedding (\uppercase\expandafter{\romannumeral3}), there are 1.4\% NDS and 2.1\% mAP performance improvements compared with the ray-aware position embedding. Moreover, our OPE outperforms the point-aware position embedding with 0.8\% NDS and 0.8\% mAP, which illustrates the effectiveness and superiority of OPE for enhancing detection performance. 

\noindent\textbf{Effectiveness of the OPE for distant objects.}
To verify the effectiveness of the OPE for distant objects, we compare different position embedding methods based on distance on the nuScenes \texttt{val} set. Specifically, we set the minimum distance thresholds between the ground truth and ego as 0, 20, and 40 meters, respectively. As shown in Table~\ref{tab:dist}, when the thresholds are set to 0, 20, and 40 meters, OPE can bring 1.4\% NDS, 1.5\% NDS, and 2.3\% NDS performance improvement, respectively. These experiments indicate that as the distance increases, the performance gain brought by OPE gradually becomes larger compared with the ray-aware position embedding of StreamPETR. Besides, when the thresholds are set to 40 meters, OPE outperforms the point-aware position embedding of 3DPPE by 1.2\% NDS performance improvement. Therefore, these experiments demonstrate the importance of OPE for distant objects.

\begin{table*}[t!]
\centering
\caption{
Ablation studies for the necessity of pixel-wise depth encoder on the nuScenes \texttt{val} set. The PDE represents the pixel-wise depth encoder.}
\setlength{\tabcolsep}{10pt}
\resizebox{1.0\linewidth}{!}{
\begin{tabular}{c|cc|ccccc}
\toprule
PDE & NDS$\uparrow$ & mAP$\uparrow$ & mATE$\downarrow$  & mASE$\downarrow$ & mAOE$\downarrow$ & mAVE$\downarrow$ & mAAE$\downarrow$ \\
\midrule
 & 59.9 & 49.8 & 0.538 & 0.258 & 0.282 & 0.221 & \textbf{0.199} \\
 \checkmark & \textbf{61.3} & \textbf{52.1} & \textbf{0.525} & \textbf{0.256} & \textbf{0.281} & \textbf{0.216} & \textbf{0.199} \\
\bottomrule
\end{tabular}
}
\label{tab:abs_pde}
\end{table*}

\noindent\textbf{Ablations of the PDE.}
We conduct ablation studies to demonstrate the necessity of pixel-wise depth encoder (PDE) for accurate object-wise depth prediction in {\ourMethod}. As shown in Table~\ref{tab:abs_pde}, when PDE is not used, we find there is a 1.4\% NDS and 2.3\% mAP performance drop. We think taking pixel-wise depth as the prior is necessary to obtain accurate object-wise depth. This experiment illustrates the necessity of PDE.

\subsection{Visualization}
\begin{figure}[t!]
\centering
    \includegraphics[width=0.99\linewidth]{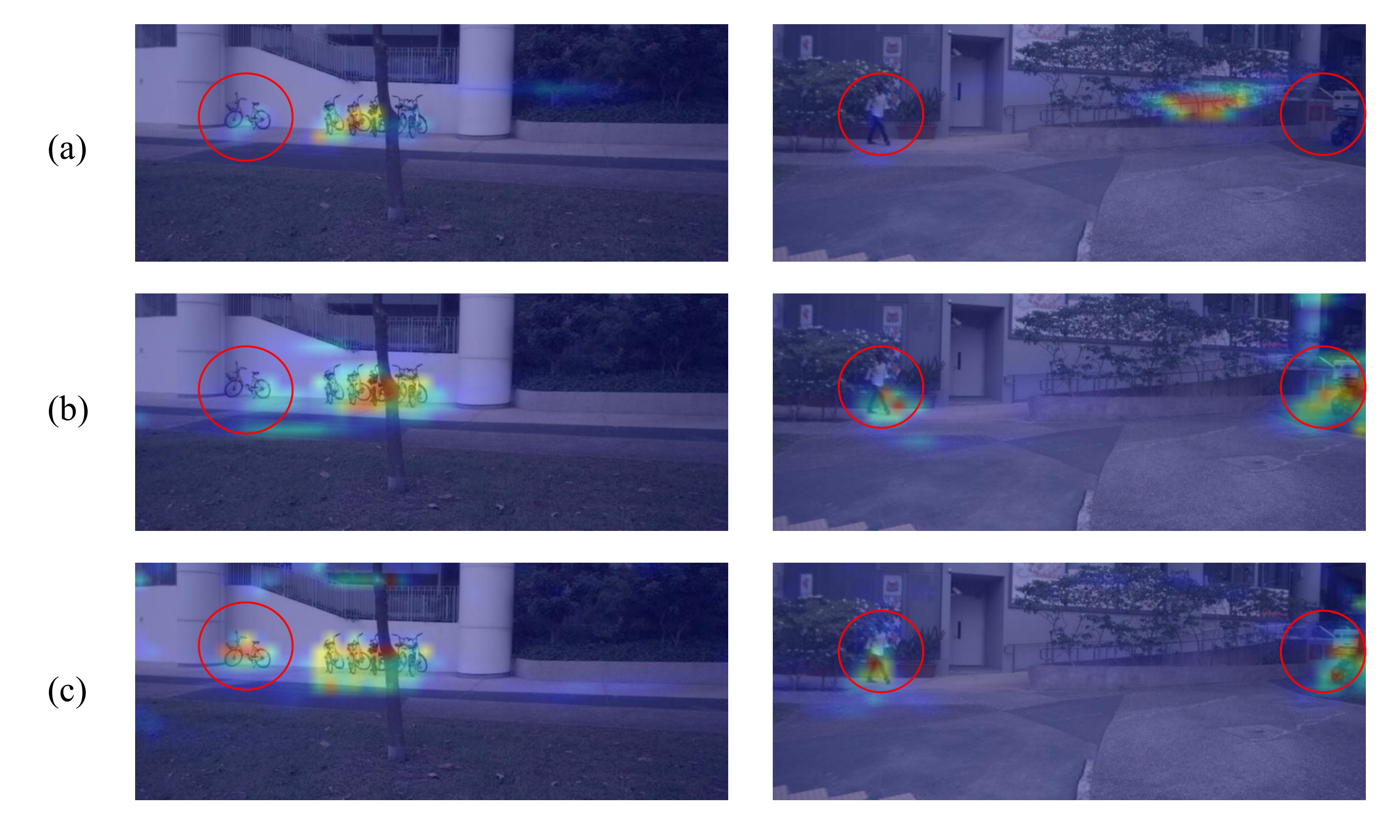}
\caption{Comparison of attention weight maps for the ray-aware position embedding of StreamPETR (a), the point-aware position embedding of 3DPPE (b), and the object-wise position embedding (c) on the nuScenes \texttt{val} set. After utilizing OPE, our {\ourMethod} can generate better attention weight maps for some hard-detected objects, which are highlighted by red circles.}
\label{fig_attn}
\end{figure}

\noindent\textbf{Attention Map Comparison.}
To explore the influence of different position embedding methods for the transformer decoder, we visualize the attention weight map of the last transformer decoder layer for the ray-aware position embedding of StreamPETR~\cite{wang2023exploring}, the point-aware position embedding of 3DPPE~\cite{shu20233dppe}, and our OPE. As shown in Figure~\ref{fig_attn}, compared with other methods, our OPE has higher attention weights for some hard-detected objects, which are highlighted by red circles. 
Furthermore, the attention weight map obtained by OPE is more focused than point-aware position embedding. 
These visualization results indicate the effectiveness and superiority of OPE compared with other position embedding methods.

\begin{figure}[t!]
\centering
    \includegraphics[width=0.99\linewidth]{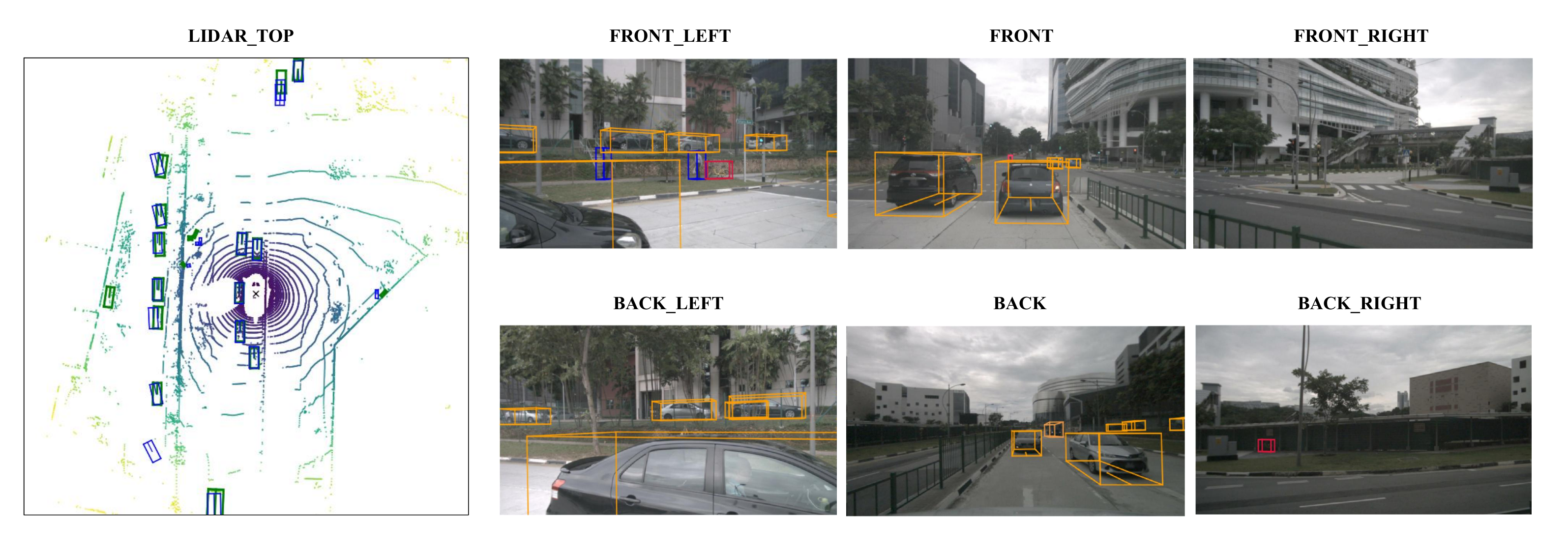}
\caption{Qualitative detection results on multi-view images and the BEV space on the nuScenes \texttt{val} set. The 3D predicted bounding boxes are drawn with different colors for different classes on multi-view images. Blue and green boxes are the prediction and ground truth boxes on the BEV space.}
\label{fig_res}
\end{figure}

\noindent\textbf{Qualitative Results.}
As shown in Figure~\ref{fig_res}, we show the qualitative detection results of {\ourMethod} on multi-view images and BEV space. We can observe that our {\ourMethod} can detect some distant objects successfully. More qualitative results are provided in the supplemental material.

\section{Conclusion}
\label{sec:conclusion}

In this paper, we have proposed a new multi-view 3D object detector named {\ourMethod}, which mainly introduces the object-wise depth information to multi-view 3D object detection for 3D object-aware feature representation by our proposed object-wise position embedding. Specifically, 
{\ourMethod} first employs an object-wise depth encoder, which combines temporal information and takes the pixel-wise depth map as a prior, to accurately estimate the object-wise depth. Then, {\ourMethod} utilizes
the proposed object-wise position embedding to encode the object-wise depth information into the transformer decoder, thereby producing 3D object-aware features for final detection.
Besides, we design the depth-aware focal loss to encourage {\ourMethod} to pay more attention to the 3D object center information. 
Extensive experiments demonstrate the effectiveness of each component of our {\ourMethod}. Furthermore, {\ourMethod} has achieved state-of-the-art detection performance on the nuScenes dataset.
Finally, we hope {\ourMethod} could further promote the research of depth in multi-view 3D object detection.


\clearpage  

%
%
\bibliographystyle{splncs04}
\bibliography{main}

\clearpage
\appendix
\section{Appendix}
\subsection{Comparison of Depth Prediction}
In this section, we report the related depth estimation metric~(L1 Error and AbsReal) for pixel-wise and object-wise depth of foreground objects. Besides, we also report these metrics for distant objects ($>$ 40m). As shown in Table~\ref{tab:depth}, we can find that the accuracy of object-wise depth prediction is better than pixel-wise. For distant objects, the difference is even more significant. These experiments indicate the object-wise depth is more easily estimated, specifically for some distant objects.

\begin{table}[h!]
\small
\centering
\vspace{-10pt}
\caption{
Comparison of pixel-wise depth prediction and object-wise depth prediction on nuScenes \texttt{val} set.}
\setlength{\tabcolsep}{5pt}
\resizebox{1.0\linewidth}{!}{
\begin{tabular}{c|c|c|c|c}
\toprule
 Depth Prediction & L1 Error (m) $\downarrow$ & AbsRel $\downarrow$ & L1 Error$_{>40}$ (m) $\downarrow$ & AbsRel$_{>40}$ $\downarrow$\\
\midrule
Pixel-wise  & 4.44 & 0.233 & 12.28 & 0.515 \\
Object-wise  & \textbf{2.20} & \textbf{0.090} & \textbf{3.88} & \textbf{0.106}\\
\bottomrule
\end{tabular}
}
\vspace{-15pt}
\label{tab:depth}
\end{table}

\subsection{Comparison of Efficiency}
In this section, we compare the running speed, computation cost, and parameter size with StreamPETR~\cite{wang2023exploring} and 3DPPE~\cite{shu20233dppe}. We adopt the same experiment setting as the ablation studies of the main paper. For a fair comparison, the running speed of all methods is evaluated on an NVIDIA GeForce RTX 3090 with a batch size of 1. As shown in Table~\ref{tab:fps}, our method brings satisfactory performance improvements under the comparable running speed, computation cost, and parameter size compared with other methods. 

\begin{table*}[h!]
\centering
\vspace{-10pt}
\caption{
Comparison of running speed, computation cost, and parameter size with other methods. For a fair comparison, the running speed of all methods is evaluated on an NVIDIA GeForce RTX 3090 with a batch size of 1.}
\setlength{\tabcolsep}{10pt}
\resizebox{1.0\linewidth}{!}{
        \begin{tabular}{c|cc|c|c|c}
        \toprule
        Method & NDS$\uparrow$ & mAP$\uparrow$ & Speed (FPS) & FLOPs (G) & Params (M)\\
        \midrule
        StreamPETR~\cite{wang2023exploring} & 59.4 & 50.3 & 3.5 & 524.6 & 73.2\\
        3DPPE~\cite{shu20233dppe} & 60.0 & 51.6 & 3.5 & 558.8 & 79.2 \\
        {\ourMethod} & 61.3 & 52.1 & 3.5 & 560.8 & 79.6 \\
        \bottomrule
    \end{tabular}
}
\label{tab:fps}
\vspace{-10pt}
\end{table*}

\begin{figure}[t!]
\centering
    \includegraphics[width=0.99\linewidth]{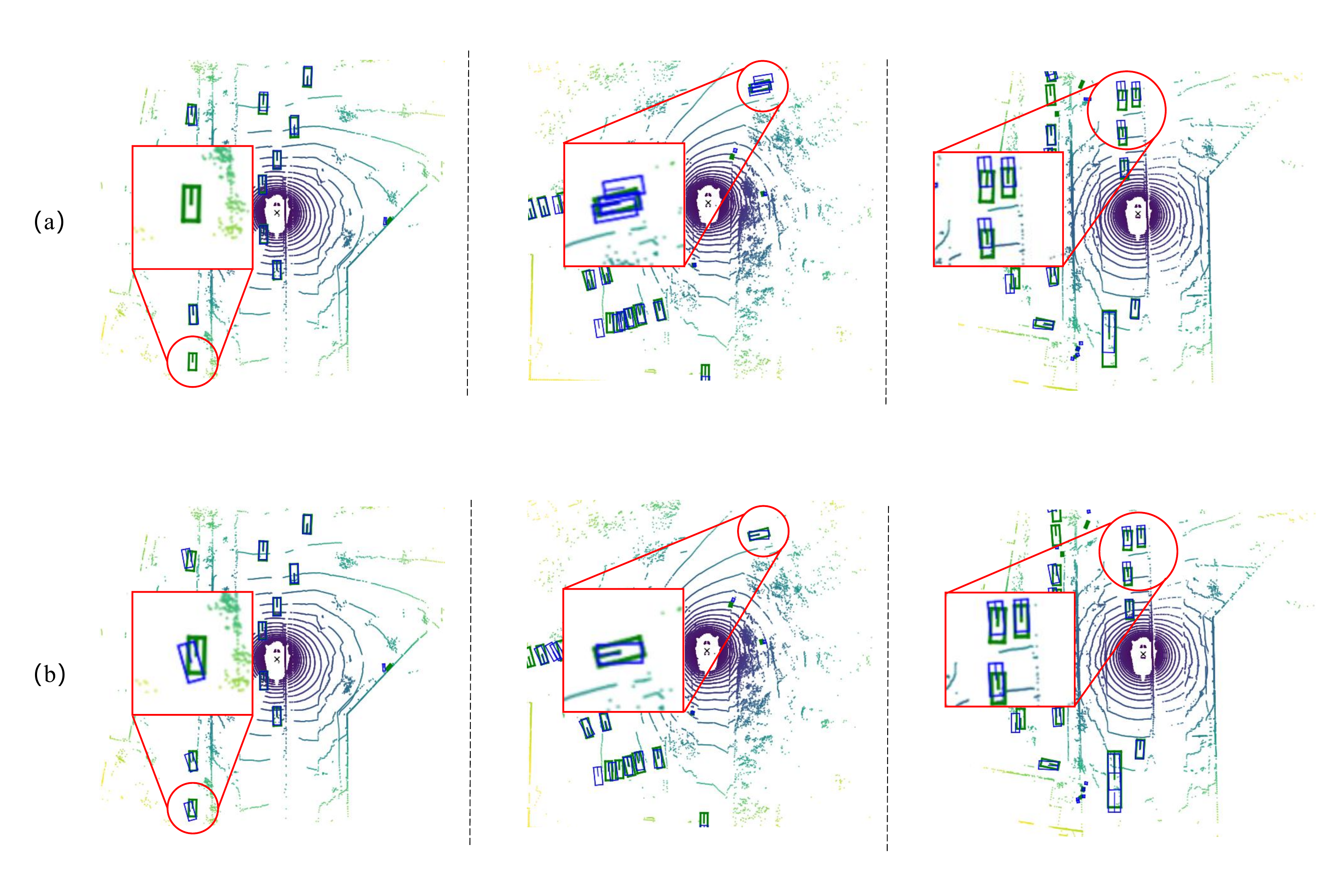}
\caption{Comparison of StreamPETR (a) and {\ourMethod} (b) on the nuScenes \texttt{val} set. Blue and green boxes are the prediction and ground truth boxes. In the first column, {\ourMethod} can successfully detect the hard-detect object. In the second column, {\ourMethod} can distinguish the false positive better. In the third column, {\ourMethod} can achieve more accurate location ability for distant objects.}
\label{fig_cmp}
\end{figure}

\subsection{More Experiments}
\label{sec:exp}
In this section, we conduct experiments with the ViT-L~\cite{fang2023eva} backbone to demonstrate the effectiveness of {\ourMethod} and conduct experiments with the single frame as input to demonstrate the generalization of {\ourMethod} on nuScenes \texttt{val} set.

\noindent\textbf{Performance of {\ourMethod} with ViT backbone.} To further verify the effectiveness of {\ourMethod}. We validate our method with the ViT-L backbone with an image size of $320\times800$ on nuScenes \texttt{val} set. For all methods, we train 24 and 48 epochs. As shown in Table~\ref{tab:vit}, when training 24 epochs, {\ourMethod} achieves 61.6\% NDS and 52.3\% mAP, outperforming StreamPETR~\cite{wang2023exploring} by 1.5\% NDS and 0.8\% mAP. When training 48 epochs, {\ourMethod} achieves 62.5\% NDS and 54.2\% mAP, outperforming StreamPETR by 1.2\% NDS and 1.3\% mAP. These experiments demonstrate the effectiveness of {\ourMethod} with a larger backbone.

\begin{figure}[t!]
\centering
    \includegraphics[width=0.99\linewidth]{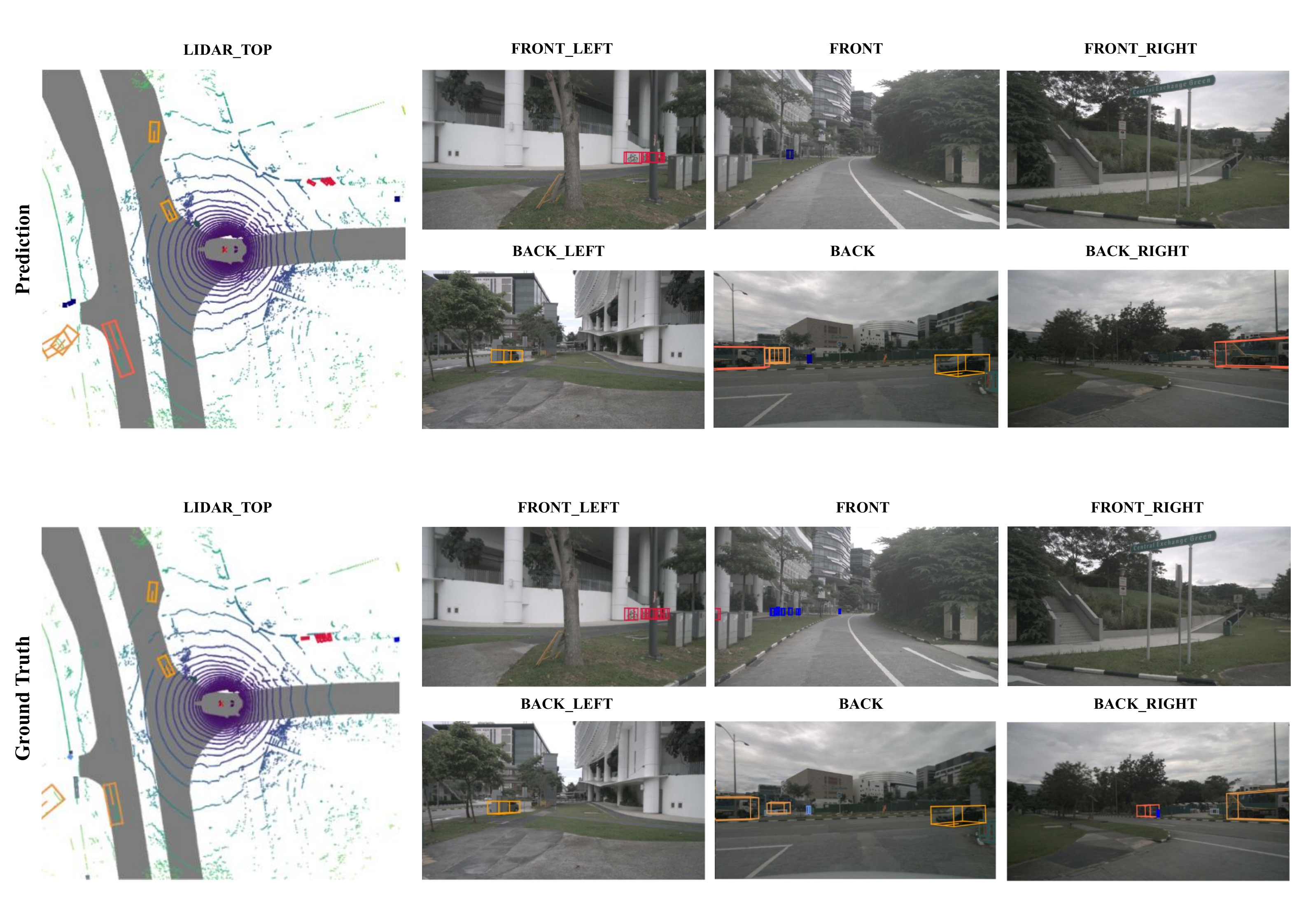}
\caption{Qualitative detection results on multi-view images and BEV space on the nuScenes \texttt{val} set. The 3D predicted bounding boxes are drawn with different colors for different classes.}
\label{fig_res}
\vspace{-10pt}
\end{figure}

\begin{table*}[t!]
\centering
\caption{
Comparison of other method with ViT-L backbone on nuScenes \texttt{val} set.
}
\resizebox{1.0\linewidth}{!}{
\begin{tabular}{l|c|c|c|cc|ccccc}
\toprule
    Method & Backbone & Input Size & Epoch & NDS$\uparrow$ & mAP$\uparrow$ & mATE$\downarrow$  & mASE$\downarrow$ & mAOE$\downarrow$ & mAVE$\downarrow$ & mAAE$\downarrow$\\
    \midrule
     StreamPETR~\cite{wang2023exploring} & ViT-L & $320\times800$ & 24 & 60.1 & 51.5 & 0.576 & \textbf{0.254} & 0.300 & 0.233 & 0.204 \\
     {\ourMethod} & ViT-L & $320\times800$ & 24 & \textbf{61.6} & \textbf{52.3} & \textbf{0.534} & 0.259 & \textbf{0.269} & \textbf{0.200} & \textbf{0.197} \\
     \midrule
     StreamPETR~\cite{wang2023exploring} & ViT-L & $320\times800$ & 48 & 61.3 & 52.9 & 0.563 & \textbf{0.253} & \textbf{0.266} & 0.230 & 0.198 \\
     {\ourMethod} & ViT-L & $320\times800$ & 48 & \textbf{62.5} & \textbf{54.2} & \textbf{0.511} & 0.258 & 0.295 & \textbf{0.205} & \textbf{0.197} \\
    \bottomrule
\end{tabular}
}
\vspace{-10pt}
\label{tab:vit}
\end{table*}

\begin{table*}[t!]
\centering
\caption{
Comparison of other method with single frame on nuScenes \texttt{val} set.
}
\resizebox{1.0\linewidth}{!}{
\begin{tabular}{l|c|c|cc|ccccc}
\toprule
    Method & Backbone & Input Size & NDS$\uparrow$ & mAP$\uparrow$ & mATE$\downarrow$  & mASE$\downarrow$ & mAOE$\downarrow$ & mAVE$\downarrow$ & mAAE$\downarrow$\\
    \midrule
     PETR~\cite{liu2022petr} & V2-99 & $900\times1600$ & 45.5 & 40.6 & 0.736 & 0.271 & 0.432 & 0.825 & 0.204 \\
     {\ourMethod} & V2-99 & $640\times1600$ & \textbf{50.3} & \textbf{44.0} & \textbf{0.648} & \textbf{0.265} & \textbf{0.394} & \textbf{0.668} & \textbf{0.193} \\
    \bottomrule
\end{tabular}
}
\label{tab:sf}
\vspace{-10pt}
\end{table*}

\noindent\textbf{Performance of {\ourMethod} with Single Frame.} We validate our method with single frames as input on nuScenes \texttt{val} set. We adopt the V2-99~\cite{wang2021fcos3d} backbone pre-trained on FCOS~\cite{wang2021fcos3d} with an image size of $640\times1600$ and 24 epochs. for comparison with PETR~\cite{liu2022petr}. As shown in Table~\ref{tab:sf}, even though we use a smaller input size, {\ourMethod} still achieves 50.3\% NDS and 43.0\% mAP, outperforming PETR by 4.8\% NDS and 3.4\% mAP. It is worth noting that we disable the temporal information of the object-wise depth encoder and keep the same training settings as PETR (\textit{e.g.}, without denoising training~\cite{li2022dn}, the weight of center loss is set to 1). These experiments demonstrate the generalization of {\ourMethod}.

\subsection{Visualization}
\label{sec:vis}
\noindent\textbf{Comparison of Qualitative Results.} To illustrate the superiority of our {\ourMethod}, we compare the qualitative detection results of StreamPETR and {\ourMethod} on nuScenes \texttt{val} set. As shown in Figure~\ref{fig_cmp}, in the first column, {\ourMethod} can successfully detect the hard-detect object. Then, in the second column, {\ourMethod} can distinguish the false positive better. Finally, in the third column, {\ourMethod} can achieve more accurate location ability for distant objects. These results illustrate the superiority of {\ourMethod}.

\noindent\textbf{Qualitative Results.} We show the qualitative detection results of {\ourMethod} in Figure~\ref{fig_res} on multi-view images and BEV space. The 3D predicted bounding boxes are drawn with different colors for different classes. The first row and second row are the prediction of {\ourMethod} and the ground truth, respectively.

\subsection{Limitations}
Our method utilizes global attention in the transformer decoder, which has larger computational costs when facing larger-scale scenarios. In the future, we plan to design effective local attention for the transformer decoder to improve the efficiency of {\ourMethod}.

\subsection{Potential Negative Impact}
{\ourMethod} improves 3D detection performance by introducing object-wise depth information. However, depth prediction usually requires more computation costs, leading to higher requirements for the hardware of autonomous driving.
\end{document}